\documentclass[
]{ceurart}

\usepackage{microtype}
\usepackage{listings}
\usepackage{amsmath,amssymb,amsfonts}
\usepackage{booktabs}
\usepackage{tabularx}
\usepackage{multirow}
\usepackage{graphicx}
\usepackage{subcaption}
\usepackage{algorithm}
\usepackage{algpseudocode}
\usepackage{hyperref}
\usepackage{enumitem}
\usepackage{siunitx}
\usepackage{xcolor}

\hypersetup{
  colorlinks=true,
  linkcolor=blue,
  citecolor=blue,
  urlcolor=blue
}

\lstdefinestyle{mystyle}{
    basicstyle=\ttfamily\small,
    breaklines=true,
    showspaces=false,                % This removes those "under-tabs" in general text
    showstringspaces=false,          % This removes them inside quotes
    columns=flexible,
    keepspaces=true                  % Keeps the alignment without the symbols
}

\newcommand{\tool}[1]{\textsc{#1}}
\newcommand{\dataset}[1]{\textsc{#1}}
\newcommand{\model}[1]{\textsc{#1}}
\newcommand{\smiles}{\texttt{SMILES}}
\newcommand{\gin}{\textsc{GIN}}
\newcommand{\gnne}{\textsc{GNNExplainer}}

\begin{document}

\copyrightyear{2022}
\copyrightclause{Copyright for this paper by its authors.
  Use permitted under Creative Commons License Attribution 4.0
  International (CC BY 4.0).}

\conference{2nd Causal Neuro-symbolic Artificial Intelligence (Causal NeSy): Toward Agentic LLMs with Neuro-Symbolic and Graph Based Reasoning}

\title{Improving Molecular Property Prediction in Small Language Models Using Graph-based Tools}

% \tnotetext[1]{You can use this document as the template for preparing your
%   publication. We recommend using the latest version of the ceurart style.}

\author[1]{Konstantinos Bougiatiotis}[%
orcid=0000-0002-1910-2758,
email=bogas.ko@iit.demokritos.gr,
]
\fnmark[1]
\cormark[1]

\address[1]{Institute of Informatics and Telecommunications, National Center for Scientific Research ``Demokritos", Athens, Greece}
%\address[2]{??}

\author[1]{Dimitrios Kelesis}[%
orcid=0000-0002-3434-2717,
]
\fnmark[1]

\author[1]{Georgios Paliouras}[%
]
\cortext[1]{Corresponding author.}
\fntext[1]{These authors contributed equally.}
\begin{abstract}
Small language models (SLMs) have shown promise for zero-shot molecular property prediction from \smiles{} strings, yet they often suffer from \emph{structural blindness} because sequence representations under-specify key graph-topological cues. We propose a modular \textit{Context-Augmented Prompting} framework that enables agentic tool use at inference time: a trained GNN expert model provides a predictive hint with confidence, and a \gnne{} extracts an instance-specific explanatory subgraph (e.g., a subgraph \smiles{} and an accompanying explanatory paragraph). We evaluate three commonly used SLMs on MUTAG and Tox21 under five prompting configurations ranging from SMILES-only to using all available tools at hand. Across two datasets, enriching prompts with graph-derived context yields substantial accuracy gains, often exceeding 25\% relative improvement and up to 74\% on Tox21. We further validate the functional relevance of the extracted motifs via a necessity-based edge-drop intervention. Despite the observed gains, a persistent gap remains to specialized GNN models, highlighting both the value and limits of text-conditioned reasoning for molecular structure.

\end{abstract}

\begin{keywords}
Molecule Property Prediction \sep 
Small Language Models (SLMs) \sep
Graph Neural Networks (GNNs) \sep
Explainability
\end{keywords}

\maketitle

\section{Introduction}
\label{sec:intro}

Molecular Property Prediction (MPP) remains a foundational challenge in the drug discovery pipeline~\cite{drug_pipe,drug_pipe1,drug_pipe2}, where the accurate classification of chemical compounds determines the efficiency of the Design-Make-Test-Analyze cycle. Traditionally, Graph Neural Networks (GNNs) have emerged as the state-of-the-art approach for these tasks due to their ability to directly model the topological structure of molecules~\cite{gnn_mols,gnn_mols1,gnn_mols2}. However, despite their predictive power, GNNs often operate as ``black boxes," providing little human-interpretable rationale for their outputs, which limits their utility in high-stakes medicinal chemistry decisions~\cite{xai_survey, xai_survey1}.

\noindent At the same time, Large Language Models (LLMs) have demonstrated a surprising capacity for zero-shot reasoning across various scientific domains. In the context of chemistry, LLMs typically process molecules via the Simplified Molecular Input Line Entry System (SMILES) \cite{smiles1}. While SMILES provides a compact string representation, it is a sequence-based abstraction that frequently fails to encode complex 2D branching and spatial relationships. Previous studies \cite{llm_mols,llm_mols1} suggest that LLMs struggle with ``structural blindness", leading to inconsistent performance when property prediction relies on specific local motifs or global connectivity.

\noindent To address these limitations and showcase the improvement that LMs may exhibit when they have access to expert-based tools, we propose a \textit{Context-Augmented Prompting} framework that leverages a GNN model to ground LM reasoning in the most influential subgraph identified by a GNN explainer. By extracting the causal structural subgraph given by a GNN expert, and use them as a ``hint'' about chemical properties, we provide Small Language Models (SLMs) with the structural context that raw SMILES strings lack. 

\noindent In this work, we evaluate three prominent SLMs; namely Llama 3.2 (3B parameters)~\cite{llama}, Qwen 2.5 (3B parameters) ~\cite{qwen2.5}, and DeepSeek (2.4B active parameters in a 16B Mixture-of-Experts architecture)~\cite{deepseek}, across two benchmark datasets, MUTAG~\cite{mutag} and Tox21~\cite{tox21}. Our empirical findings indicate that incorporating structured contextual information alongside baseline SMILES prompts can lead to meaningful performance improvements (over $25\%$ relative improvement in almost all cases). However, the effectiveness of such explanations appears to depend strongly on the underlying language model architecture, with smaller-parameter models showing limited or inconsistent gains on certain tasks. Additionally, although GNN-augmented SLMs benefit from the integration of graph-derived hints, they do not match the performance of specialized GNN architectures, highlighting persistent challenges in fully capturing structural information through text-based reasoning alone.\\
The main contributions of this work are summarized as follows:
\begin{itemize}
    \item \textbf{A Context-Augmented Prompting Framework:} We propose a strategy that bridges the gap between graph-based structural learning and sequence-based SLM reasoning by incorporating GNN-derived hints.
    \item \textbf{Quantification of Explanation Efficacy:} We analyze the impact of the GNN-derived signals on SLM performance, showing that while such context can substantially enhance predictive accuracy, the gains are dataset- and model-dependent, and a measurable gap remains between text-augmented SLMs and specialized GNN architectures.
   \item \textbf{Comparative SLM Benchmark:} We provide an empirical evaluation of three state-of-the-art SLMs on molecular classification, highlighting the varying degrees of ``structural blindness" inherent in different model architectures.
\end{itemize}

\section{Related Work}
\label{sec:related_work}

\subsection{Structural Representation Learning in Chemistry}
The evolution of Molecular Property Prediction (MPP) has transitioned from handcrafted molecular descriptors and fingerprints, such as ECFP \cite{Rogers2010}, to deep representation learning. Graph Neural Networks (GNNs), including Graph Convolutional Networks (GCN) \cite{gcn} and Graph Isomorphism Networks (GIN) \cite{Xu2019}, are widely adopted for molecular representation learning, by treating molecules as topological graphs. While these models excel at capturing local and global chemical environments, their internal representations remain non-symbolic, making it difficult for practitioners to extract actionable chemical insights.

\subsection{Explainability in Graph Neural Networks}
To address the opacity of GNNs, various post-hoc explanation methods have been proposed. Tools such as GNNExplainer \cite{Ying2019} and PGExplainer \cite{Luo2020} identify the most influential subgraphs (motifs) contributing to a specific prediction. However, these explanations are typically provided as importance weights over edges or nodes. Transforming these numerical importance scores into human-intelligible reasoning remains a challenge, which this work addresses by using SLMs as a linguistic interface.

\subsection{LLMs for Molecular Reasoning}
Recent advancements have seen LLMs applied to chemistry via SMILES string processing. Models like Galactica \cite{Taylor2022} and MolT5 \cite{Edwards2022} demonstrated that pre-training on large scientific corpora enables basic property prediction. However, research into general-purpose models like Llama 3 and GPT-4 has revealed a ``SMILES-to-structure" gap; where these models often fail to reliably reconstruct or reason over the underlying molecular graph topology \cite{chembert}. Our work builds on this by testing if explicit structural ``hints" can mitigate this sequential bias.

\subsection{Hybrid GNN-SLM Integration}
Emerging research has explored the alignment of graph encoders with language models, often through contrastive learning or projection layers, such works include MolCA \cite{Molca} or MolLM \cite{MolLM}. Unlike these computationally expensive architectural integrations, our approach focuses on \textit{Context-Augmented Prompting}. This method preserves the frozen weights of state-of-the-art SLMs and investigates the extent to which GNN-derived explanatory signals, representing the model-identified most influential substructures, can help mitigate the modality gap. Note that this strategy is both more accessible and modular for practical deployment.

% NOTE: Sections 1-2 assumed already present in your draft.
% -------------------------------------------------------------------
\section{Methodology}
\label{sec:method}

\subsection{Problem Setting}
We study binary molecular property prediction from a textual representation.
Each molecule is provided as a \smiles{} string $s$, and the target label is
$y \in \{0,1\}$ indicating whether the molecule is toxic/mutagenic under the dataset definition.
Our goal is to improve \emph{zero-shot} performance of SLMs on this task by enabling
\emph{agentic tool use}: at inference time, the SLM can query a learned graph expert and explanation tool to obtain
structured context that is otherwise absent from the \smiles{} sequence.

\subsection{Context-Augmented Prompting Framework}

\begin{figure}[h]
    \centering
    \includegraphics[width=\textwidth]{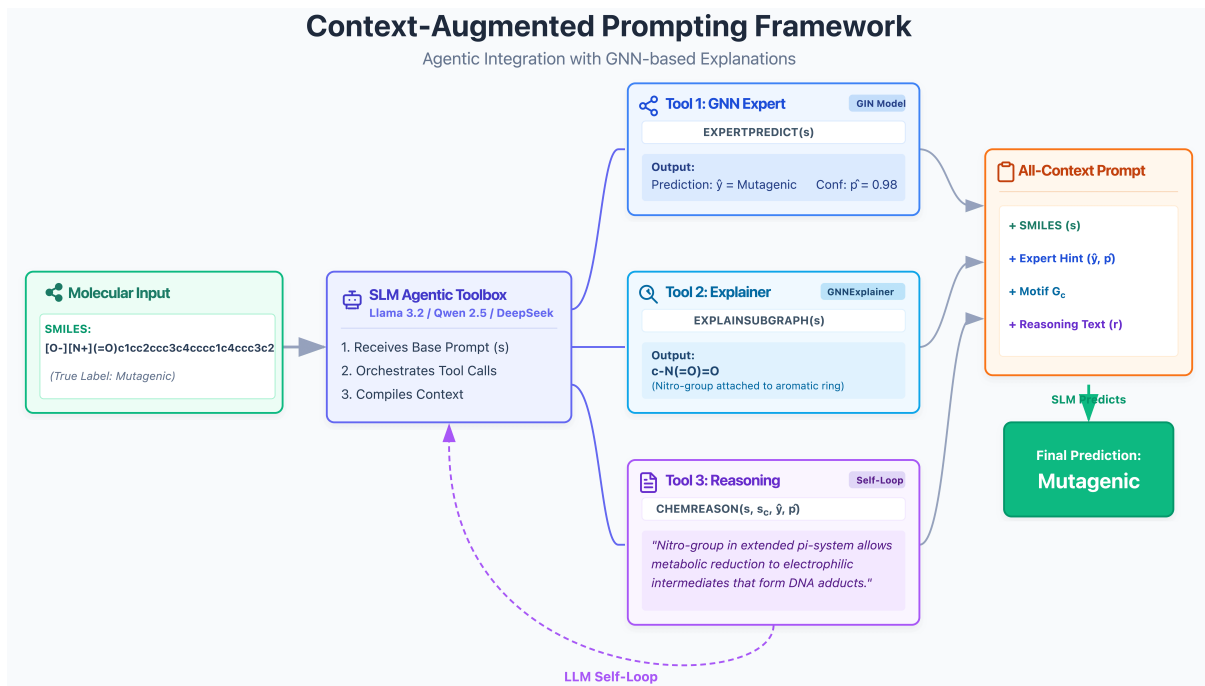}
    \caption{The proposed agentic loop. Given the query SMILES, the SLM invokes one or more of the available tools to interact with pre-trained GNN expert, to generate related context and augment the original prompt, to provide assistance for the final inference.}
    \label{fig:agentic_loop}
\end{figure}

As shown in Figure \ref{fig:agentic_loop}, our framework adopts an agentic architecture where an SLM serves as the reasoning controller, utilizing tools to interact with a GNN expert.

\noindent First, let a molecule  represented by the \smiles{} $s$ correspond to a molecular graph $G=(V,E,\mathbf{X})$ with atoms $V$, bonds $E$,
and node features $\mathbf{X}$ (e.g., atom type, valence, aromaticity). We assume that there is pre-trained GNN (expert) model available to us\footnote{Practically, we train a \gin{} model
$f_{\theta}(G)\rightarrow (\hat{y}, \hat{p})$ on the training split of each dataset, with $G$ a molecule graph,  $\hat{y}\in\{0,1\}$ and $\hat{p}\in[0,1]$
is the model confidence.}. We model inference as a single-step agent loop: given an input \smiles{} string $s$, the SLM can call tools that return
(1) expert predictions and confidence ($\hat{y},\hat{p}$), (2) the determined  explanatory subgraph ($G_c$), and (3) a natural-language rationale derived from the subgraph ($r$). Specifically, we define three primary tools that the SLM can invoke to enhance its zero-shot reasoning:

% We define three primary tools that the SLM can invoke to enhance its zero-shot reasoning.

\begin{itemize}[leftmargin=1.2em]
  \item \tool{ExpertPredict}$(s)$: converts $s \mapsto G$, returns $(\hat{y},\hat{p})$ from the pre-trained \gin{}.
  \item \tool{ExplainSubgraph}$(s)$: returns an explanatory subgraph $G_c$ (in the form of the subgraph \smiles{} $s_c$) using \gnne{}~\cite{Ying2019} over the same \gin{}.
  \item \tool{ChemReason}$(s, s_c, \hat{y}, \hat{p})$: using a short chemistry-oriented explanation templated prompt, we instruct the SLM to reason about why the motif $s_c$ might support toxicity/mutagenicity. In practice we implement this as an instruction block and ask the SLM itself to generate the rationale; conceptually it is a tool interface for a reasoning module.
\end{itemize}

\noindent To avoid introducing external knowledge bias from larger teacher models, each SLM model itself is used to generate the rationale (i.e., ChemReason), ensuring that the performance gains are strictly a product of the model's own reasoning conditioned on GNN-derived evidence.

\noindent  Regarding \tool{ExplainSubgraph}, we extract a per-instance explanatory subgraph via
\gnne{} as an edge-importance mask $\mathbf{m}\in[0,1]^{|E|}$. We then define the \emph{candidate causal subgraph}:
\begin{equation*}
G_{c} = (V_c, E_c), \quad E_c = \{ e \in E : m_e \ge \tau \},
\end{equation*}
where the threshold $\tau$ is chosen dynamically for each module, so that $E_c$ contains only edges at the top $10\%$ of the edge importance score. That is, the designated subgraph $G_c$ contains only the top-$10\%$ most important edges of the graph, as identified by GNNExplainer. To make $G_c$ consumable by text-only SLMs, we convert it to a \smiles{}
string (denoted $s_c$).

\subsection{Prompt Configurations}
Using the above tools, we evaluate five prompt configurations that correspond to progressively richer tool outputs. The variants used are shown below (the exact text of these different prompt configurations can be seen in Appendix~\ref{sec:prompts}):

\begin{enumerate}[leftmargin=1.2em]
  \item \textbf{SMILES (Baseline).} Input is only the \smiles{} string $s$.
  \item \textbf{SMILES + SUBGRAPH.} Input includes $s$ and the extracted subgraph \smiles{} $s_c$.
  \item \textbf{SMILES + HINT.} Input includes $s$ and the GNN expert output $(\hat{y},\hat{p})$ as a ``field expert hint'' (i.e., the GNN label prediction and confidence in $\%$) .
  \item \textbf{SMILES + REASONING.} Input includes $s$ and a short natural-language rationale about the motif's relevance $r$ (generated by the SLM when given $s$ and $s_c$).
  \item \textbf{ALL CONTEXT.} Input includes $s$, $(\hat{y},\hat{p})$, $s_c$ and the rationale $r$.
\end{enumerate}

\section{Experiments and Discussion}
\label{sec:exp}

\subsection{Experimental Setup}

We evaluate our framework on two standard molecular classification benchmarks: MUTAG~\cite{mutag} and a random subset of 1,000 samples from Tox21~\cite{tox21}. 
\dataset{MUTAG} contains 188 nitroaromatic and heteroaromatic compounds labeled for mutagenicity, serving as a small benchmark for graph-based molecular classification. 
\dataset{Tox21} is a multi-label toxicity benchmark that spans multiple biological targets; we sample 1,000 molecules and convert the task to binary classification by marking a molecule as positive if it has \emph{any} positive toxicity label. For each dataset, we use a 70/30 train/test split. Importantly, the split is used \emph{only} to train the \gin{} expert.
The SLMs are evaluated in a \emph{zero-shot} manner on the test portion only. The general characteristics of the datasets are shown in Table~\ref{tab:dataset_characteristics}.

\begin{table}[htbp]
\centering
\caption{Dataset characteristics.}
\label{tab:dataset_characteristics}
\begin{tabular}{lrrrr}
\toprule
\textbf{Dataset} & \textbf{\# Molecules/Graphs} & \textbf{Avg. Nodes} & \textbf{Avg. Edges} & \textbf{Positive Ratio (\%)} \\
\midrule
MUTAG        & 188  & 18.03 & 39.80 & 67 \\
Tox21 & 1000 & 18.37 & 38.99 & 37 \\
\bottomrule
\end{tabular}
\end{table}

\noindent For the GNN expert, we adopt a common \gin{} setup across datasets to reduce confounding: 3 message-passing layers, hidden dimension 64, a learning rate of $2\cdot 10^{-3}$ and trained for up to 100 epochs with early stopping. We hold out 10\% of the training split for validation and use patience 10 for early stopping.

\subsection{Small Language Models}
\noindent We evaluate three SLMs spanning different architectural and capacity regimes: \model{DeepSeek} (Lite-Code), a Mixture-of-Experts model with 16B total parameters and 2.4B active parameters; \model{Qwen 2.5} (Code), with 3B parameters; and \model{Llama 3.2}, also with 3B parameters. \model{DeepSeek} is optimized for code-centric instruction following and structured reasoning.\model{Qwen 2.5} is likewise tuned for step-wise generation and robust formatting, making it well suited to evidence-conditioned decision prompts. \model{Llama 3.2} serves as a general-purpose baseline, enabling comparison between code-tuned models and a more broadly trained conversational model under identical zero-shot prompting. All models are evaluated in zero-shot mode on the test splits of the datasets, using fixed prompts for each configuration. We report classification \textbf{accuracy} (\%) for each prompt configuration.

\subsection{Main Results}
Table~\ref{tab:main} summarizes the accuracy over the different prompt configurations and the two datasets.

\begin{table}[htbp]
\centering
\caption{Model accuracy (in \%) across different prompt configurations.  The highest accuracy per SLM and dataset is highlighted in bold. The ``Improvement'' row indicates the relative percentage gain of the ALL CONTEXT configuration compared to the baseline SMILES prompt.}
\label{tab:main}
\begin{tabular}{lcccccc}
\toprule
\multirow{2}{*}{\textbf{Configuration}} & \multicolumn{3}{c}{\textbf{MUTAG}} & \multicolumn{3}{c}{\textbf{Tox21}} \\
\cmidrule(lr){2-4} \cmidrule(lr){5-7}
& \textbf{DeepSeek} & \textbf{Qwen 2.5} & \textbf{Llama 3.2} & \textbf{DeepSeek} & \textbf{Qwen 2.5} & \textbf{Llama 3.2} \\
\midrule
SMILES              & 56.14 & 55.36 & \textbf{47.37} & 38.50 & 47.50 & 48.50 \\
SMILES + SUBGRAPH   & 54.39 & 66.07 & 43.86 & 45.50 & 47.24 & 55.00 \\
SMILES + HINT       & 42.11 & 49.12 & 43.86 & \textbf{67.00} & 67.50 & \textbf{64.00} \\
SMILES + REASONING  & 66.67 & 49.12 & 40.35 & 55.50 & 51.50 & 61.50 \\
ALL CONTEXT         & \textbf{75.44} & \textbf{70.18} & \textbf{47.37} & \textbf{67.00} & \textbf{68.50} & 63.50 \\
\midrule
\textit{Improvement over SMILES} & \textcolor{green!60!black}{$\uparrow$ 34.38\%} & \textcolor{green!60!black}{$\uparrow$ 26.77\%} & 0.00 & \textcolor{green!60!black}{$\uparrow$ 74.03\%} & \textcolor{green!60!black}{$\uparrow$ 44.21\%} & \textcolor{green!60!black}{$\uparrow$ 30.93\%} \\
\midrule
\textbf{GNN Expert} & \multicolumn{3}{c}{84.21} & \multicolumn{3}{c}{71.00} \\
\bottomrule
\end{tabular}
\end{table}

% \begin{table}[t]
% \centering
% \small
% \setlength{\tabcolsep}{6pt}
% \begin{tabular}{lcccccc}
% \toprule
% \multirow{2}{*}{Configuration} &
% \multicolumn{3}{c}{\dataset{MUTAG}} &
% \multicolumn{3}{c}{\dataset{Tox21} (1k)} \\
% \cmidrule(lr){2-4}\cmidrule(lr){5-7}
% & DeepSeek & Qwen 2.5 & Llama 3.2 & DeepSeek & Qwen 2.5 & Llama 3.2 \\
% \midrule
% SMILES & 56.14 & 55.36 & 47.37 & 38.50 & 47.50 & 48.50 \\
% SMILES + SUBGRAPH & 54.39 & 66.07 & 43.86 & 45.50 & 47.24 & 55.00 \\
% SMILES + HINT & 42.11 & 49.12 & 43.86 & 67.00 & 67.50 & 64.00 \\
% SMILES + REASONING & 66.67 & 49.12 & 40.35 & 55.50 & 51.50 & 61.50 \\
% ALL CONTEXT & \textbf{75.44} & \textbf{70.18} & \textbf{47.37} & \textbf{67.00} & \textbf{68.50} & \textbf{63.50} \\
% \midrule
% Improvement over SMILES & $\uparrow$34.38\% & $\uparrow$26.77\% & 0.00\% & $\uparrow$74.03\% & $\uparrow$44.21\% & $\uparrow$30.93\% \\
% \midrule
% GNN Expert & \multicolumn{3}{c}{84.21} & \multicolumn{3}{c}{71.00} \\
% \bottomrule
% \end{tabular}
% \caption{Accuracy (\%) across prompt configurations. Best per SLM/dataset in bold.}
% \label{tab:main}
% \end{table}

\paragraph{Observation 1: Tool augmentation yields large gains on the harder setting.}
On \dataset{Tox21}, the baseline SMILES-only prompting is low for DeepSeek (38.50\%) and moderate for Qwen/Llama (47--49\%).
Adding tools closes much of this gap, with ALL CONTEXT reaching 67.00\% (DeepSeek) and 68.50\% (Qwen), representing large relative gains
(74.03\% and 44.21\% relative improvements over baseline, respectively).
This is consistent with the claim that raw \smiles{} induces ``structural blindness'' while graph-derived context restores missing relational information.

\paragraph{Observation 2: Model capacity/competence mediates benefit from structured context.}
The strongest and most consistent improvements are observed for DeepSeek and Qwen.
Llama 3.2 shows mixed behavior: some single-tool variants help on \dataset{Tox21} (e.g., SUBGRAPH and HINT),
but the ALL CONTEXT improvement is smaller and in \dataset{MUTAG} does not improve beyond baseline, suggesting that smaller-scale general-purpose models may suffer from ``information distraction". 
%or an inability to synthesize heterogeneous structural hints when the baseline chemical reasoning is already weak.
This suggests that benefiting from tool outputs requires the SLM to (i) integrate heterogeneous evidence and (ii) maintain instruction-following under longer contexts.

\paragraph{Observation 3: Single-tool signals can be unstable; combined context is more robust.}
Across datasets, isolated signals sometimes degrade performance (e.g., MUTAG with SMILES+HINT for DeepSeek).
This is expected in an agentic setting: expert hints may be miscalibrated, and subgraphs may be hard to interpret without narrative scaffolding.
ALL CONTEXT often recovers and improves, indicating complementary effects:
\emph{(a) expert prediction provides a prior}, \emph{(b) motif provides structural evidence}, \emph{(c) reasoning promotes semantic alignment}
between text and graph evidence.

\paragraph{Observation 4: The gap to specialized graph models remains.}
Even with ALL CONTEXT, SLMs do not consistently surpass the \gin{} expert (84.21\% on MUTAG; 71.00\% on Tox21).
This highlights a boundary of purely text-conditioned reasoning: graph inductive bias and representation learning remain crucial for molecular structure.

\subsection{Complementary Experiments and a Qualitative Case Study}\label{sec:necessity}

\begin{figure}[t]
\centering
\includegraphics[width=0.78\linewidth]{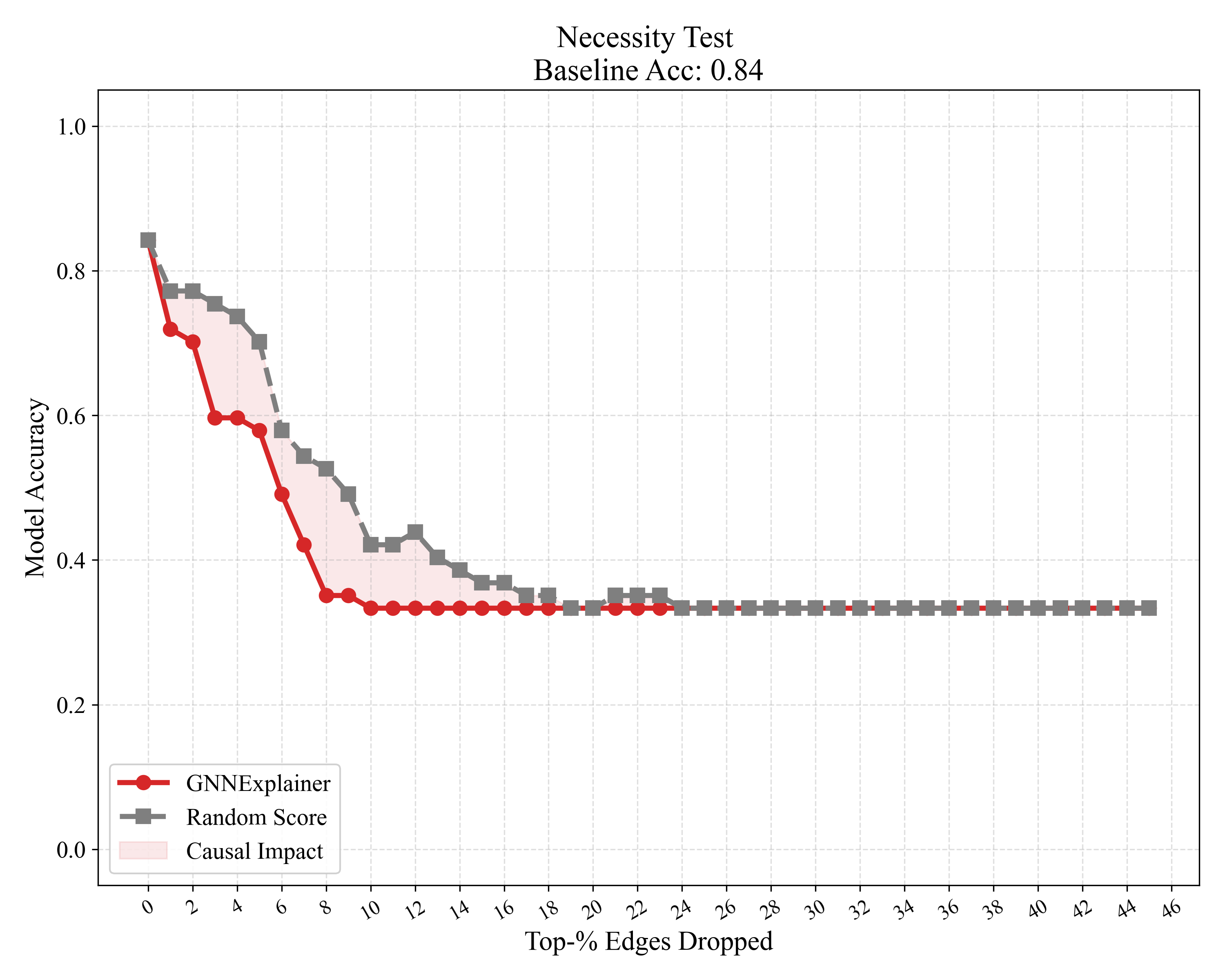}
\caption{Necessity test, showing the \gin{} accuracy as edges are removed in explainer-importance order (red) vs random removal (grey).
A larger separation indicates that the extracted motif is closer to a necessary substructure for the expert prediction.}
\label{fig:necessity}
\end{figure}

In this section, we first present a supplementary experiment examining the necessity of the highlighted subgraphs identified by GNNExplainer, thereby supporting the importance of the selection. We then illustrate, through a sample-based case study, a scenario in which the SLM, using the reasoning tool, can disregard the incorrect input provided by the GNNExpert. 
% Finally, we discuss the limitations of the current study.

\noindent To assess whether the extracted motif is merely correlational or behaves like a \emph{necessary} substructure for the expert's prediction,
we perform a targeted perturbation test on the expert model:
we progressively remove edges from $G$ either: (i) in decreasing importance order according to \gnne{} (\textbf{GNNExplainer}),
or (ii) at random (\textbf{Random Score}), and measure the expert accuracy as a function of the number of removed edges.
A strong separation between these curves indicates that explainer-ranked edges concentrate predictive signal and removing them is a stronger intervention,
consistent with a necessity-style causal diagnostic. 

\noindent Figure~\ref{fig:necessity} reports the necessity test as described. We can see that removing a small fraction of explainer-ranked edges produces a much steeper accuracy drop than removing the same number of random edges.
Concretely, the curve indicates that dropping roughly \(\sim 8\%\) of top explainer edges yields a degradation comparable to dropping roughly \(\sim 25\%\) of randomly chosen edges, illustrating that explainer-selected edges concentrate decision-critical information. The plateau after $25$\% for both scenarios is the result of most graphs becoming disconnected so the message-passing regime of the GNN expert breaks down.

% \noindent\textbf{Interpretation (workshop alignment).}
% This experiment operationalizes a lightweight causal notion (necessity under intervention) in a neuro-symbolic pipeline:
% the explanation is not merely an attention map, but an intervention target that can be perturbed to validate its functional role.
% This connects to \emph{counterfactual reasoning} (``what if the motif were absent?'') and \emph{verifiable agent behavior} (the agent can present evidence that survives intervention tests).

\begin{figure}[t]
\centering
\includegraphics[width=0.95\linewidth]{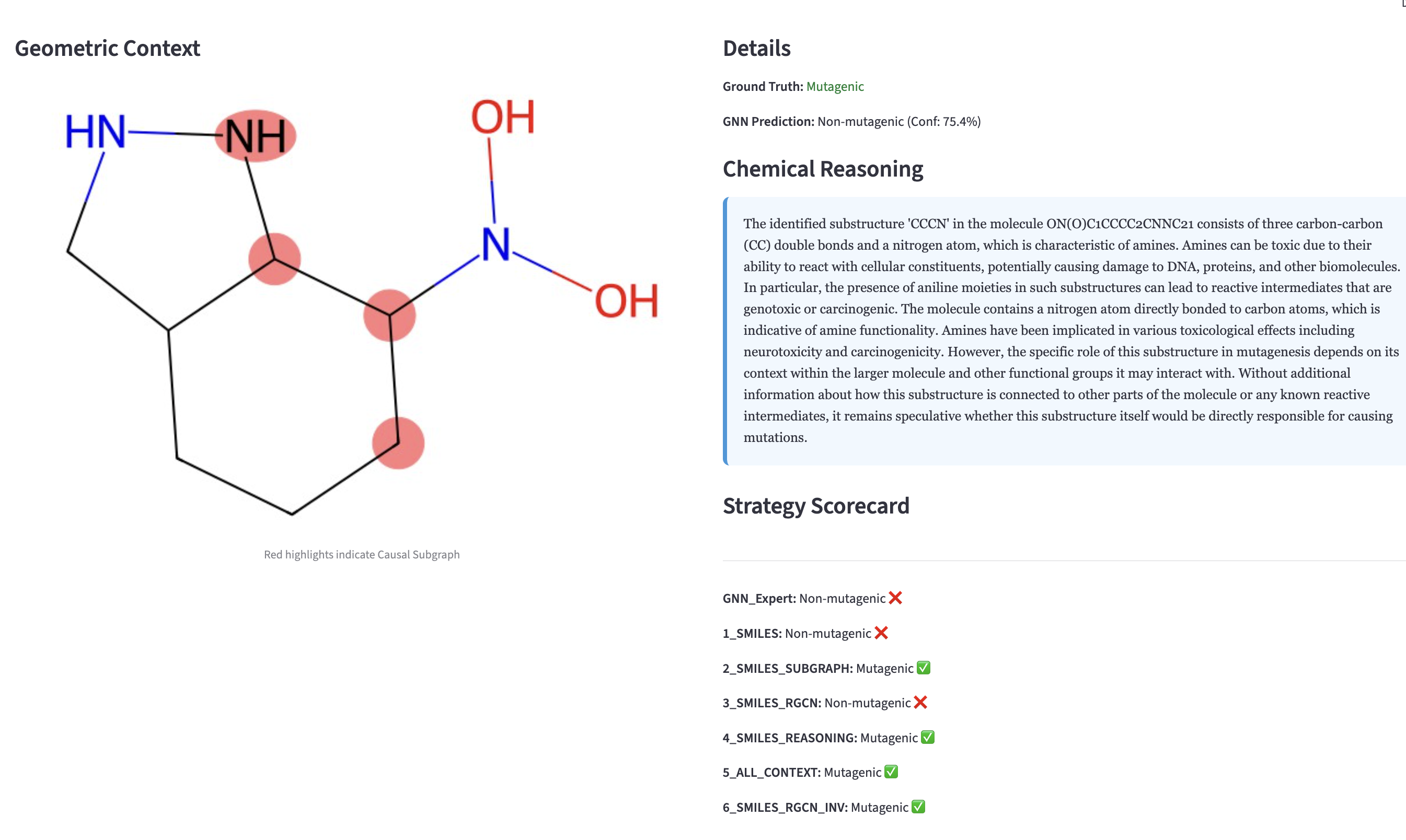}
\caption{Example showing that motif-based reasoning can improve the final prediction (and, in some cases,
correct the GNN expert) by using the subgraph as structured evidence and reason on its importance.}
\label{fig:case}
\end{figure}

\noindent Moving on to the case study, Figure~\ref{fig:case} illustrates a case where providing the motif and asking the SLM to reason about its chemical implications
helps produce a correct label even when the \gin{} expert is wrong.
This behavior is important for agentic deployments: tools are not always oracle-correct, and a reliable agent should be able to
\emph{cross-check} tool outputs using additional evidence and domain constraints. The case study suggests a two-level governance pattern: the graph expert provides a strong inductive bias,
while the SLM serves as a \emph{semantic controller} that can (i) interpret explanations, (ii) identify inconsistencies,
and (iii) decide whether to accept or ignore the expert output. This resembles \emph{human-in-the-loop causal evaluation},
except here the ``human'' role is approximated by language-based reasoning conditioned on explicit evidence.

\section{Conclusion}
\label{sec:concl}
We presented a simple, modular approach to improving zero-shot molecular property prediction with small language models by enabling tool use.
A trained GNN expert provides predictive hints and a \gnne{}-derived explanatory motif, which together serve as structured, auditable context that mitigates
the structural limitations of \smiles{}-only prompting.
Across \dataset{MUTAG} and a subset of \dataset{Tox21}, we find that enriching prompts with expert confidence, subgraph evidence, and motif-based reasoning
can yield substantial gains, particularly on the more challenging \dataset{Tox21} setting, while still leaving a gap to specialized graph models.
We further validated that explainer-selected edges behave like necessary structure for the expert under an intervention-style edge-drop test,
supporting a causal grounding perspective.
Overall, these results position tool-augmented SLMs as lightweight agentic systems that combine neural prediction with symbolic graph evidence,
offering a practical step toward causally grounded neuro-symbolic agents in scientific domains.

\section*{Declaration on Generative AI}
During the preparation of this work, the authors used GPT-5, Gemini 3, and Grammarly for the purposes of: Drafting content, Grammar and spelling check, Paraphrase and reword, Improve writing style.\\ 
After using these tools/services, the authors reviewed and edited the content as needed and take full responsibility for the publication’s content. 

\bibliography{biblio}

% \clearpage
\appendix

\section{Prompts Used in Configurations}\label{sec:prompts}

The following table contains the prompts used for the different scenarios as explained in the Methodology Section~\ref{sec:method}. To make sure the SLMs generate valid labels, we validate their output using Pydantic~\footnote{\url{https://docs.pydantic.dev/latest/}} templates, allowing only for binary predictions when running an inference setup. All SLMs are locally hosted using an Ollama~\footnote{\url{https://ollama.com/}} server.

\begin{table}[ht]
\begin{small}

\centering
\caption{Different prompts used with SLMs. The ``Common'' and ``Task'' prompts are used at the start and end of each variant. The term ``Toxic'' is substituted with ``Mutagenic'' when running on the MUTAG dataset. }\label{tab:zero_shot_prompts}
\begin{tabularx}{\linewidth}{cX}
\hline
\textbf{Variant} & \textbf{Prompt}\\
\hline
Common & {\ttfamily You are an expert toxicologist. Classify the molecule as Toxic (True) or Non-toxic (False).} \\
Task & \texttt{Predict toxicity based on this SMILES: \{full\_smiles\}.} \\
\hline
SMILES & Common + Task \\
\hline
SMILES + SUBGRAPH & Common + \texttt{Critical Subgraph: \{sub\_smiles\}.} + Task \\
\hline
SMILES + HINT   & Common + \texttt{An expert model prediction is that it may be \{gnn\_pred\} (confidence: \{gnn\_conf\} \%).} + Task \\
\hline
SMILES + REASONING  & Common + \texttt{A possible chemical reasoning for the molecule's label is: \{chem\_reason\}.} + Task \\
\hline
ALL CONTEXT  & Common + All of the above + Task \\
\hline

\hline
\end{tabularx}
\end{small}
\end{table}

\noindent Regarding the \tool{ChemReason} tool that generates the reasoning paragraph for each subgraph, we use the following prompt template for each SLM:

\begin{lstlisting}[style=mystyle]
"""    
You are a computational toxicologist.

Task: Given the molecule {full_smiles} and the subgraph {sub_smiles},
write ONE concise paragraph explaining the potential chemical impact of the substructure.
If the substructure contains known toxicophores, explain why they are dangerous.
If the substructure appears generic or does NOT contain known toxic features,
explicitly state that it may not be toxic or that its role is unclear. 
Do not force a toxic explanation if one does not exist.
"""
\end{lstlisting}

\end{document}